\documentclass[10pt, a4paper]{article}
\usepackage{lrec2022} 
\usepackage{multibib}
\newcites{languageresource}{Language Resources}
\usepackage{graphicx}
\usepackage{tabularx}
\usepackage{soul}
\usepackage{float}
\usepackage{multirow}
\usepackage{amsmath}
\usepackage{tablefootnote}
\usepackage{subcaption}
\usepackage{tablefootnote}

\usepackage{titlesec}
\titleformat{\section}{\normalfont\large\bfseries\center}{\thesection.}{1em}{}
\titleformat{\subsection}{\normalfont\SmallTitleFont\bfseries\raggedright}{\thesubsection.}{1em}{}
\titleformat{\subsubsection}{\normalfont\normalsize\bfseries\raggedright}{\thesubsubsection.}{1em}{}
\renewcommand\thesection{\arabic{section}}
\renewcommand\thesubsection{\thesection.\arabic{subsection}}
\renewcommand\thesubsubsection{\thesubsection.\arabic{subsubsection}}

\usepackage{epstopdf}
\usepackage[utf8]{inputenc}

\usepackage{hyperref}
\usepackage{xstring}
\usepackage[demo]{}

\usepackage{color}

\newcounter{daggerfootnote}

\let\svthefootnote\thefootnote
\newcommand\freefootnote[1]{%
  \let\thefootnote\relax%
  \footnotetext{#1}%
  \let\thefootnote\svthefootnote%
}

\title{An Evaluation Framework for Legal Document Summarization}

\name{
\renewcommand*{\thefootnote}{\fnsymbol{footnote}}
Ankan Mullick$^{\spadesuit\dagger}$~~~~
Abhilash Nandy$^{\spadesuit\diamondsuit\dagger}$~~~~
Manav Nitin Kapadnis$^{\spadesuit\dagger}$~~~~\vspace{4pt}\\
\hspace{-40pt}
\textbf{\large{Sohan Patnaik}}$^{\spadesuit}$\hspace{15pt}
\textbf{\large{R Raghav}}$^{\spadesuit}$\hspace{15pt}\textbf{\large{Roshni Kar}}$^{\spadesuit}$\\ \vspace{-5pt}
\\
$^\spadesuit$Indian Institute of Technology Kharagpur ~
$^{\diamondsuit}$ L3S Research Center, Leibniz Universität Hannover
}

\address{\{ankanm, nandyabhilash\}@kgpian.iitkgp.ac.in,\\\{iammanavk, sohanpatnaik106, rraghav5600, roshnikar\}@iitkgp.ac.in}

\abstract{
A law practitioner has to go through numerous lengthy legal case proceedings for their practices of various categories, such as land dispute, corruption, etc. Hence, it is important to summarize these documents, and ensure that summaries contain phrases with intent matching the category of the case. To the best of our knowledge, there is no evaluation metric that evaluates a summary based on its intent. We propose an automated intent-based summarization metric, which shows a better agreement with human evaluation as compared to other automated metrics like BLEU, ROUGE-L etc. in terms of human satisfaction. We also curate a dataset by annotating intent phrases in legal documents, and show a proof of concept as to how this system can be automated. Additionally, all the code and data to generate reproducible results is available on \href{https://github.com/manavkapadnis/LegalEvaluation_LREC2022}{Github}. 
 \\ \newline 
 }

\begin{document}

\maketitleabstract

\freefootnote{$^{\dagger}$Authors contributed equally}

\section{Introduction}

Summarization could be extractive, where the summary has spans from the original text or abstractive, where the summary is generated using the original text. \cite{eval_nlg}, \cite{eval_qa} list various metrics to evaluate summarization. BLEU~\cite{bleu}, METEOR~\cite{meteor}, ROUGE~\cite{lin-2004-rouge} etc. are context-free metrics, which work well for extractive summarization, while SMS~\cite{clark2019sentence}, BERTScore~\cite{bert-score} etc. are context-based metrics, which work well for abstractive summarization. Automatic summarization of legal documents~\cite{bhattacharya2019} is required because - (1) Average length of an any Court Judgement is as high as $4,500$ words (for example - Indian Supreme Court Judgments) (2) A law practitioner has to go through all contents of previous legal proceedings manually (3) Hiring Legal experts to summarize legal documents is expensive and very time consuming. However, while evaluating the quality of summaries, existing metrics fail when evaluating the amount of intent in the original text that is captured by the summary (intent here refers to the intention latent in a piece of text. e.g. (a)`Accused No. 1 Balwan Singh (appellant in Criminal Appeal No. 727 of 2015), on 22nd January, 2007, at evening time, was talking with the other accused regarding preparation to kill' - in this sentence, the phrase `preparation to kill' depicts the intent of Murder (b) `In the case in hand,
robbed articles were found to be kept concealed at a place within knowledge of the
applicant/accused No.1, and therefore, he is presumed to be one of the decoit involved in the decoity at the house of the first Aarti Palkar' - in this sentence, the phrases `robbed articles were found to be kept concealed' and `involved in the decoity' depict the intent of Robbery). To tackle this problem, we propose a novel evaluation metric that takes the help of intent phrases annotated using legal case documents, such that, the intent of these phrases matches with the category of the case. We use this proposed metric (and other metrics) to evaluate unsupervised summarization methods on legal documents and compare this with human evaluation. Another contribution is the curation of a dataset that consists of $101$ legal documents spanning four categories of intents - Corruption, Land Dispute, Murder, and Robbery, along with a list of annotated intent phrases per document. Additionally, we come up with a framework, showing that the annotation of intent phrases and classification of documents into categories of intents can be automated. You can test our methods using this demo website\footnote{Demo website: \url{https://bit.ly/demoLREC2022}}. 
\section{Related Work}
\label{related_work}
\if{0}
In a country like India having highest population, abundant volume of information in the legal domain gets generated in every passing day.India having 24 high courts and 600 district courts,there are huge number of legal cases pending across different courts of the country.In view of the fact that most court judgments are predominantly lengthy documents, headers generated act as summary for these judgments. However the synopsis process involves veteran legal experts. Therefore, the whole process becomes time-consuming task involving considerable human engagement and brain storming.\newline
Automation of the summarization of Indian legal case documents helps in identifying the important paragraphs from these documents and thus reducing the human effort.Text summarization techniques help in reducing documents with a system in order to make a summary that keeps important information of the original documents and thus in generating headers.With incessant increase amount of electronic documents, use of such systems become progressively pertinent and inevitable.\newline
There has been significant research performed leading to explore different techniques of text summarization and to evaluate these techniques. Legal text summarization is way different from general text summarization and therfore requires its own study.Therefore,  a comprehensive overview of the methods and techniques used in summarization with a focus on legal text summarization.


\noindent \textbf{Unsupervised Summarization:} Unsupervised Approaches use semantic and analytical signals
from the text to point out significant sentences for summarization. Some approaches are : (1) \textbf{CaseSummarizer}~\cite{casesummarizer}: It is specific to the legal domain, that builds on existing methods to present an interface with scalable summary text, lists of entities and abbreviations, and a significance heat map of the text. This approach consists of : pre-processing, scoring of sentence relevance, and domain processing. (2) \textbf{BERT Extractive Summarizer}~\cite{bert_extractive}: BERT \cite{bert} is modified to yield sentence embeddings by sending tokenized sentences to BERT. The sentence vectors are passed through different layers to get document level features. Finally, the summary prediction is compared to ground truth, and the loss is calculated to train both summarization layers and BERT.

\noindent \textbf{Unsupervised Summarization:} Supervised approaches take in documents and ground truth summaries, and use sentence features (e.g., facts of the case, background etc.) to filter good candidates for inclusion in summary. Some approaches are: (1) \textbf{Graphical Model}~\cite{graphical_model}: A CRF model is trained using lexical and syntactic features to classify parts of the document into different categories such as `facts', `arguments' etc. Then, a K-mixture model is used to rank sentences, and a summary of the desired length is the output. (2) \textbf{LetSum}~\cite{letsum}: LetSum extracts important sentences by connecting the topical structure in the document and certainty of contentious themes of sentences in the judgment. (3) Longformer Encoder Decoder (LED) \cite{longformer}: LED \cite{led} is a Longformer \cite{longformer} variant that supports long document generative sequence-to-sequence tasks, making it simple to process documents of thousands of tokens or longer. The length of the input document can be upto $16,384$ tokens.
\fi

\noindent \textbf{Unsupervised Summarization:} Unsupervised Approaches (\cite{unsupvermaextractive},\cite{casesummarizer},\cite{bert_extractive})  use semantic and analytical signals
from the text to point out significant sentences for summarization. CaseSummarizer~\cite{casesummarizer} is specific to the legal domain that builds on existing methods to present an interface with scalable summary text, lists of entities and abbreviations, and a significance heat map of the text. 
BERT Extractive Summarizer~\cite{bert_extractive} yields sentence embeddings by sending tokenized sentences to BERT~\cite{bert} which are passed through hidden layers to get document level features. Finally, the summary prediction is compared to the ground truth.

\noindent \textbf{Supervised Summarization:} Supervised approaches (\cite{see2017get},\cite{graphical_model},\cite{letsum},\cite{led},\cite{legal-led}) take in documents and ground truth summaries, and use sentence features (e.g., facts of the case, background etc.) to filter good candidates for inclusion in summary. Graphical CRF Model~\cite{graphical_model} is trained using lexical and syntactic features to classify parts of the document into different categories such as `facts', `arguments' etc. Then, a K-mixture model is used to rank sentences, and a summary of the desired length is the output. LetSum~\cite{letsum} extracts important sentences by connecting the topical structure in the document and certainty of contentious themes of sentences in the judgment. Longformer Encoder Decoder (LED)~\cite{longformer} and \cite{led} supports long document generative sequence-to-sequence tasks, making it simple to process documents of thousands of tokens or longer. 

But none of these works are judged against the intent of the case. Our proposed intent metrics for legal document summarization shows better relevance and human judgmental scores. 

\section{Proposed Evaluation Metric}
\if{0}
calculated as the cosine similarity between the encoded embedding vectors of $P_i$ and $O_j$ using a sentence transformer \cite{reimers-2019-sentence-bert} 
\fi 

\if{0}
\begin{equation}
s_{ij} = \frac{P_i.O_j}{|P_i|\times|O_j|}
\end{equation}
\fi

\if{0}
\begin{equation}
    s_{ij}= 
\begin{cases}
    1,& \text{if } \exists k, P_i = O_j\mathbf{[}k:k + \text{length}(P_i)\mathbf{]}\\
    0,              & \text{otherwise}
\end{cases}
\end{equation}
\fi

\if{0}
\begin{equation}
    P_{int} = \frac{\sum\limits_{j=1}^{N}\mathbf{1}_{\big[\max\limits_{i=1..M}s_{ij} > \tau\big]}}{N} \label{p_eqn}
\end{equation}

\begin{equation}
    R_{int} = \frac{\sum\limits_{i=1}^{M}\mathbf{1}_{\big[\max\limits_{j=1..N}s_{ij} > \tau\big]}}{M} \label{r_eqn}
\end{equation}
\fi

\if{0}
\noindent\begin{minipage}{.35\linewidth}
\begin{equation}
    P_{int} = \frac{\sum\limits_{j=1}^{N}\mathbf{1}_{\big[\sum\limits_{i=1}^{M}s_{ij} > 0\big]}}{N} \label{p_eqn}
\end{equation}
\end{minipage}%
\begin{minipage}{.355\linewidth}
\begin{equation}
    R_{int} = \frac{\sum\limits_{i=1}^{M}\mathbf{1}_{\big[\sum\limits_{j=1}^{N}s_{ij} > 0\big]}}{M} \label{r_eqn}
\end{equation}
\end{minipage}%
\begin{minipage}{.35\linewidth}
\begin{equation}
    F1_{int} = \frac{2.P_{int}.R_{int}}{P_{int}+R_{int}} \label{f1_eqn}
\end{equation}
\end{minipage}
\fi
\if{0}
where $\tau$ is a heuristically determined threshold for similarity score, and\fi

We introduce an intent-based F1-Score and Human Score (related to Spearman Rank Correlation) metric for evaluation of a summary, referred to as `Intent Metric' hereon. We report the average Intent Metric over all documents. Let us define 'closePair' as a pair of intent phrase and a sentence from the summary, such that, the intent phrase is contained in the sentence. The fraction of sentences in the summary that form a 'closePair' with atleast one intent phrase gives precision. 
Similarly, fraction of intent phrases that form a 'closePair' with atleast one sentence from the summary gives recall. 
Finally, Intent Metric 
is the F1 Score obtained from the precision and recall values. 
Given a document, the corresponding set $P$ of $M$ intent phrases and output summary $O$ consisting of $N$ sentences, a similarity score $s_{ij}$ between $i_{th}$ intent phrase ($P_i$) and $j_{th}$ sentence in the summary ($O_j$) is $1$ if $P_i$ is a phrase contained in $O_j$ and $0$ otherwise, $\forall i \in \{1,2,..,M\}, \forall j \in \{1,2,..,N\}$. 
Mathematically,

\if{0}
\begin{equation}
s_{ij} = \frac{P_i.O_j}{|P_i|\times|O_j|}
\end{equation}
\fi

\begin{equation}
    s_{ij}= 
\begin{cases}
    1,& \text{if } \exists k, P_i = O_j\mathbf{[}k:k + \text{length}(P_i)\mathbf{]}\\
    0,              & \text{otherwise}
\end{cases}
\end{equation}

$P_{int}$, $R_{int}$ and $F1_{int}$ are calculated in the following manner -
\noindent\begin{minipage}{.5\linewidth}
\begin{equation}
    P_{int} = \frac{\sum\limits_{j=1}^{N}\mathbf{1}_{\big[\sum\limits_{i=1}^{M}s_{ij} > 0\big]}}{N} \label{p_eqn}
\end{equation}
\end{minipage}%
\begin{minipage}{.5\linewidth}
\begin{equation}
    R_{int} = \frac{\sum\limits_{i=1}^{M}\mathbf{1}_{\big[\sum\limits_{j=1}^{N}s_{ij} > 0\big]}}{M} \label{r_eqn}
\end{equation}
\end{minipage}%
\begin{equation}
    F1_{int} = \frac{2.P_{int}.R_{int}}{P_{int}+R_{int}} \label{f1_eqn}
\end{equation}

Additionally, we also measure Precision, Recall, and F1 Score Metrics for evaluation of Slot, Intent, and Document Classification in Section~\ref{summ_methods}.
\if{0}
Note that, we use this kind of a soft contextual match instead of an exact match while calculating similarity so that this metric is applicable for both extractive as well as abstractive summarization.
\fi
\section{Dataset Description} 
\label{sec:dataset}

$5000$ legal documents are scraped from CommonLII~\footnote{\url{http://www.commonlii.org/resources/221.html}} using `selenium' python package. $101$ documents belonging to the categories of Corruption, Murder, Land Dispute, and Robbery are randomly sampled from this larger set. 

In case of Australian dataset (abbreviated as "AD"), we downloaded the Legal Case Reports Dataset\footnote{\url{https://archive.ics.uci.edu/ml/datasets/Legal+Case+Reports}} from the UCI Machine Learning repository. The annotators then manually annotate randomly taken $59$ relevant documents belonging to Corruption, Murder, Land Dispute, and Robbery categories.

Intent phrases are annotated for each document in the following manner - 
\begin{enumerate}
    \item \textbf{Initial filtering:} $2$ annotators filter out sentences that convey an intent matching the category of the document at hand.
    \item \textbf{Intent Phrase annotation} $2$ other annotators then extract a span from each sentence, so as to exclude any details do not contribute to the intent (such as name of the person, date of incident etc.), and only include the words expressing corresponding intent. The resulting spans are the intent phrases. Overall Inter-annotator agreement (Cohen $\kappa$) is 0.79.
\end{enumerate}

Table~\ref{tab:datasets_stats} shows the statistics of both the datasets, describing the number of documents, average length of documents, and intent phrases for each of the $4$ intent categories. The documents on Robbery and Land Dispute are roughly longer than those on Murder and Corruption.\footnote{We have used NLTK and Spacy for data pre-processing.}

\begin{table}[h]
\centering
\resizebox{\columnwidth}{!}{\begin{tabular}{|l|cc|cc|cc|cc|}
\hline
\textbf{Category} &
  \multicolumn{2}{c|}{\textbf{\begin{tabular}[c]{@{}c@{}}No. of\\  docs\end{tabular}}} &
  \multicolumn{2}{c|}{\textbf{\begin{tabular}[c]{@{}c@{}}Avg. no. of\\ words/doc\end{tabular}}} &
  \multicolumn{2}{c|}{\textbf{\begin{tabular}[c]{@{}c@{}}Avg. no. of\\ sentences/doc\end{tabular}}} &
  \multicolumn{2}{c|}{\textbf{\begin{tabular}[c]{@{}c@{}}Avg. no. of\\  words/intent \\phrase\end{tabular}}} \\
  \hline
                                     & ID & AD & ID                        & AD    & ID & AD  & \begin{tabular}[c]{@{}c@{}}ID\end{tabular} & AD \\
                                     \hline
Corruption   & 19 & 15 & 2542 & 4613  & 197  & 264 & 6  & 6  \\
Land Dispute & 27 & 14 & 2461  & 11508 & 196  & 579 & 5  & 6  \\
Murder       & 32 & 15 & 1560 & 3008  & 149  & 183 & 6  & 5  \\
Robbery      & 23 & 15 & 1907  & 7123  & 162  & 449 & 4  & 5 \\
\hline
\end{tabular}}
\caption{Statistics for each category in both the datasets (ID - Indian-Data, AD - Australian-Data). The numbers are rounded to the nearest integer.}
\label{tab:datasets_stats}
\end{table}




\section{Experiments and Results}
\label{summ_methods}

\noindent \textbf{Competing baselines of Summarization Methods\footnote{We used Pytorch and Tensorflow for model implementation}}: The following summarization Methods (discussed in Section~\ref{related_work}) are used in an unsupervised setting - 

\begin{enumerate}
 \item \textbf{Graphical Model}~\cite{graphical_model} - Model trained on annotated data released in \cite{bhattacharya2019} is used for inference. \item \textbf{LetSum}~\cite{letsum} - The process suggested in \cite{bhattacharya2019} is used for inference.
\item \textbf{Legal-Longformer Encoder Decoder (Legal-LED)}~\cite{legal-led} - Longformer Encoder Decoder~\cite{longformer} trained on sec litigation releases~\cite{sec-liti} is used for inference. 
\item \textbf{BERT Extractive Summarizer}~\cite{bert_extractive}
\end{enumerate}
    
\noindent \textbf{Document Classification}\\

\begin{table}[H]
\renewcommand\thetable{3}
\centering
\resizebox{\columnwidth}{!}{\begin{tabular}{|l|c c|c c|}
\hline
\textbf{Model Name} &
  \multicolumn{2}{c|}{\textbf{Accuracy}} &
  \multicolumn{2}{c|}{\textbf{\begin{tabular}[c|]{@{}c@{}}Macro F1\end{tabular}}} \\
  \hline
& ID & AD   & ID & AD           \\
\hline
Logistic Regression & 0.62  & 0.50  & 0.38 & 0.47          \\
SVM & 0.62  & 0.50 & 0.38  & 0.42          \\
AdaBoost & \textbf{0.81}  & 0.67  & \textbf{0.78} & 0.58          \\
BERT    & 0.70  & \textbf{0.75} & 0.69 & 0.64          \\
RoBERTa & 0.75  & 0.67          & 0.70   & 0.60          \\
ALBERT  & 0.70  & 0.67          & 0.69 & 0.60          \\
DeBERTa & 0.75  & 0.67          & 0.71  & 0.60          \\
LEGAL-BERT & \textbf{0.80}  & \textbf{0.75} & \textbf{0.79}  & \textbf{0.73}          \\
LEGAL-RoBERTa & 0.67  & \textbf{0.75} & 0.65 & 0.64       \\
\hline
\end{tabular}}
\caption{Results of Document Classification.}
\label{tab:baseline_results}
\end{table}
Recent developments show that, Transformer \cite{vaswani2017attention} based pre-trained language models like BERT \cite{devlin2019bert}, RoBERTa \cite{liu2019roberta}, ALBERT \cite{lan2020albert}, and DeBERTa \cite{he2021deberta}, have proven to be very successful in learning robust context-based representations of lexicons and applying these to achieve state of the art performance on a variety of downstream tasks such as document classification in our case.

We implemented different machine learning and transformer-based models mentioned in Table \ref{tab:baseline_results}.
Furthermore, We also tried domain-specific LEGAL-BERT \cite{chalkidis2020legalbert} and LEGAL-RoBERTa \footnote{\url{https://huggingface.co/saibo/legal-roberta-base}} which were pre-trained on large scale legal corpora which in turn led to much better scores than their counterparts pre-trained on general corpora. 

 
 We observe from Table \ref{tab:baseline_results} that boosting algorithms such as AdaBoost \cite{adaboost_paper} and domain pre-trained transformer models such as LEGAL-BERT outperforms all the other models in terms of Accuracy and Macro F1-score in both the ID and AD datasets.
 

All of the transformer models were implemented using sliding window attention \cite{sliding_window}, since the document length for all the documents is greater than the transformer maximum token size. They were trained with a sliding window ratio of 20\% over three epochs with learning rate and batch size set at $2\times10^{-5}$ and 32 respectively. The documents in the dataset are randomly split into train, validation and test sets in the ratio of 6:2:2. The machine learning models were implemented on the TF-IDF features extracted from of the document text.


\vspace{2pt}
\noindent \textbf{Intent Classification Using JointBERT}
\label{auto_extr}\\

\begin{table}[H]
\centering
\resizebox{\columnwidth}{!}{
\begin{tabular}{|l|c c|c c|}
\hline
\textbf{Model Name} &
  \multicolumn{2}{c|}{\textbf{Accuracy}} &
  \multicolumn{2}{c|}{\textbf{\begin{tabular}[c]{@{}c@{}}Macro F1\end{tabular}}} \\
  \hline
& ID & AD            & ID                        & AD            \\     \hline
JointBERT   & 0.89  & \textbf{0.85} & 0.88 & \textbf{0.84} \\
JointDistilBERT  & \textbf{0.95}  & 0.70 & \textbf{0.95} & 0.69 \\
JointALBERT & 0.89  & 0.71  & 0.87 & 0.68         \\
\hline
\end{tabular}}
\caption{Results on Intent classification.}
\label{tab:coarse_grain_results}
\end{table}

We used Joint-BERT~\cite{jointbert} model on both the 'Indian-Data' as well as 'Australian-Data' for the task of intent classification between the classes of `Corruption', Land Dispute', `Robbery' and `Murder'. 
The dataset is prepared in the following manner - 
Since there is a majority of sentences that have no intent phrase,
only sentences containing an intent phrase, the one before that, and the one after that are used for training to mitigate class imbalance.
Each sentence with an intent phrase has a target intent. The dataset is further randomly split into train (60\%), validation (20\%) and test sets (20\%).


The different variations of JointBERT model perform reasonably well on the intent classification task for both the datasets, as seen from Table~\ref{tab:coarse_grain_results}.
\subsection{Evaluation using automated metrics}
\label{auto_metrics}

The following baseline metrics are used for comparison with the proposed metric - 

\noindent \textbf{BLEU~\cite{bleu}}: It computes the
    number of n-grams in the predicted and reference summary. Overall score is found by taking the geometric mean of scores for n from $1$ to $4$.
    
    \noindent \textbf{METEOR~\cite{meteor}}: It is F-measure based metric 
    operating on unigrams by aligning and mapping each token in the predicted summary to a token in a
    reference summary.
    
    \noindent \textbf{ROUGE-L \cite{lin-2004-rouge}}: It is an F-measure metric based on the longest common subsequence (LCS) between the reference and generated summary.
    
    \noindent \textbf{Sentence and Word Mover Similarity (S+WMS) \cite{clark2019sentence}}: A linear programming solution measures the distance a predicted summary's embedding has to be moved to match the reference, and the similarity metric is calculated. 
    
    \noindent \textbf{BERTScore~\cite{bert-score}}: It obtains BERT~\cite{bert} representations of each word in the
    predicted and reference summaries.
    Finally, a modified F1 score (weighted using
    inverse-document-frequency values) is found.

\begin{figure*}[t]
	\centering
	\subfloat[On `Indian-Data'(0.3)]{%
		\centering \includegraphics[width=0.33\textwidth]{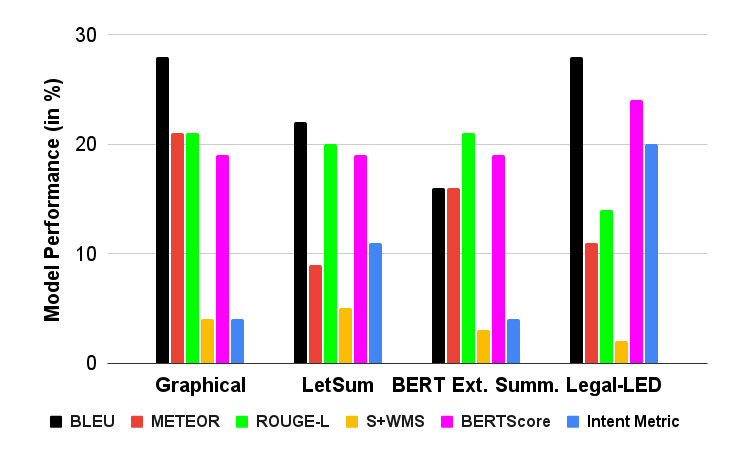} \label{id_0.3}}
	\subfloat[On `Indian-Data'(0.5)]{%
		\centering \includegraphics[width=0.33\textwidth]{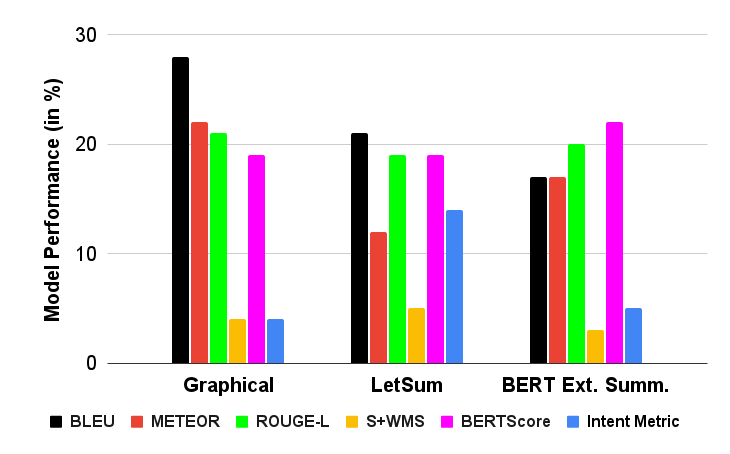} \label{id_0.5}}
	\subfloat[On `Indian-Data'(0.7)]{%
		\centering \includegraphics[width=0.33\textwidth]{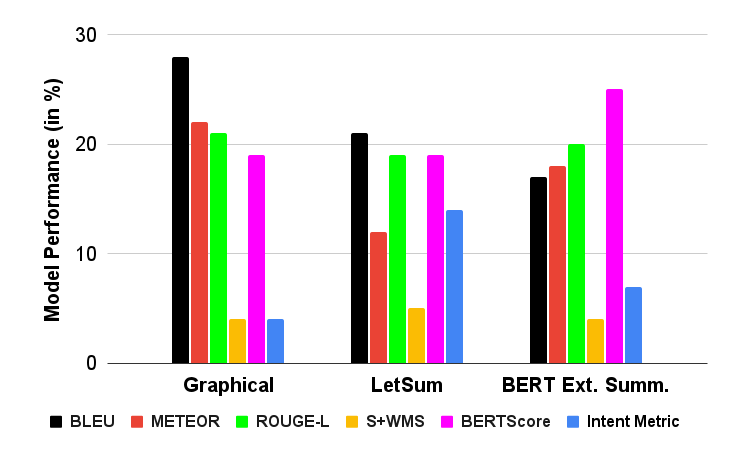} \label{id_0.7}}\\
	\subfloat[On `Australian-Data'(0.3)]{%
		\centering \includegraphics[width=0.33\textwidth]{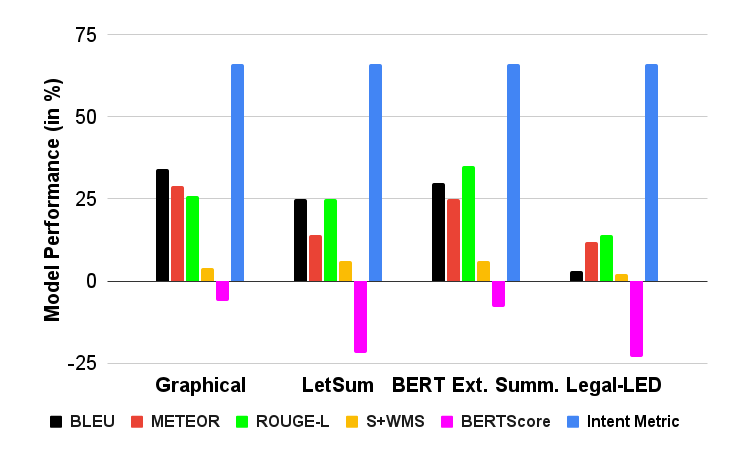} \label{ad_0.3}}
	\subfloat[On `Australian-Data'(0.5)]{%
		\centering \includegraphics[width=0.33\textwidth]{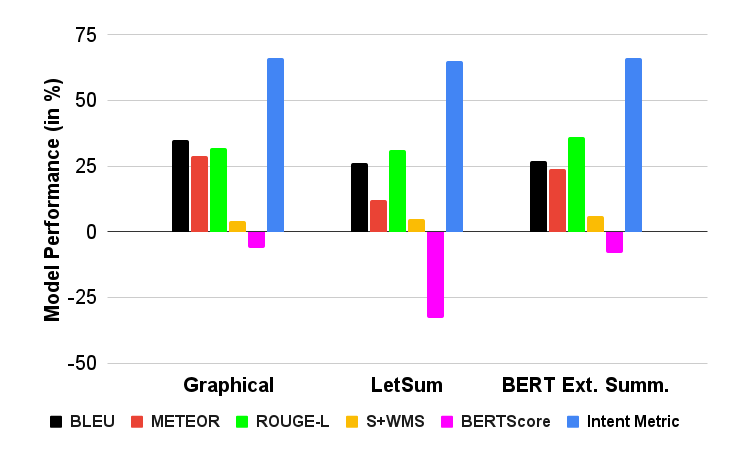} \label{ad_0.5}}
	\subfloat[On `Australian-Data'(0.7)]{%
		\centering \includegraphics[width=0.33\textwidth]{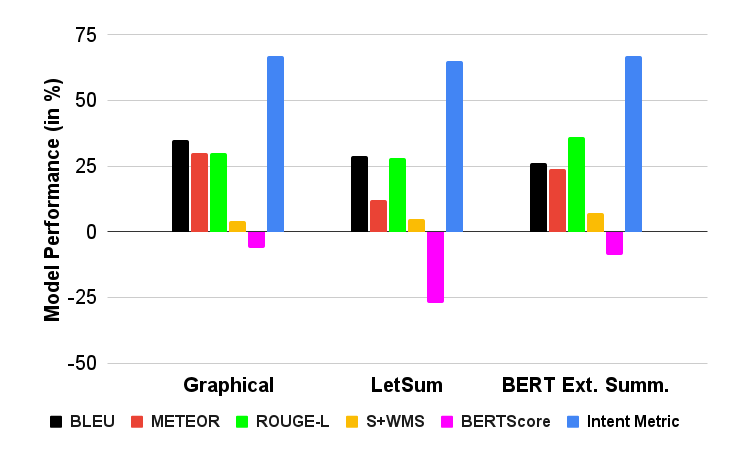} \label{ad_0.7}}\\

	\caption{Model evaluation results on different ratios mentioned in parenthesis on ID and AD. Ratio here is (length of summary/length of original text)}
	\label{fig:summarization_plots}
\end{figure*}

Fig.~\ref{fig:summarization_plots} plots the evaluation metrics for the two datasets and different lengths of summary as a fraction of the original document length (fractions are $0.3,0.5,0.7$). In some cases, BERTScore is negative as BERTScore ranges from $-1$ to $1$. Also, \emph{Legal-LED consumes more than 95\% of GPU memory when the summary length is $50\%$ and $70\%$ of the original text, and hence, could not be reported.} The scores do not depend significantly on the summary length as a fraction of the input. However, we cannot conclude if one metric is better than the other, as every metric has its own way of quantifying the summary quality.  comparing the three models - (1) Graphical Model tends to preform the best for lexical metrics such as BLEU, METEOR, ROUGE-L. (2) BERT Extractive Summarizer gives the best BERTScore, as is expected. (3) Legal-LED performs better on `Indian Data' compared to `Australian Data'. (4) In case of `Indian Data', LetSum performs the best as per Intent Metric and S+WMS, while in case of `Australian Data', all models perform almost equally well w.r.t these metrics. (5) Given a dataset, Intent Metric significantly varies across different summarization methods, which makes Intent Metric human-readable. To compare the quality of metrics, we see how well they correlate with human judgement in Section~\ref{humaneval}.

\begin{figure}[H]
	\centering
	\subfloat[On `Indian-Data']{%
		\centering \includegraphics[width=0.5\columnwidth]{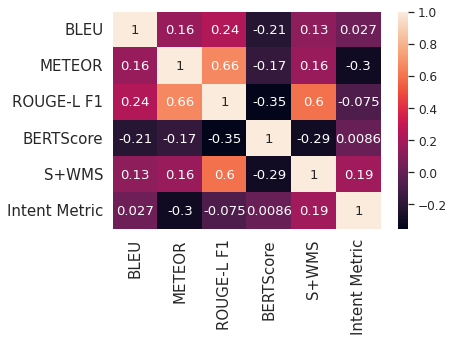} \label{corr_1}}
	\subfloat[On `Australian-Data']{%
		\centering \includegraphics[width=0.5\columnwidth]{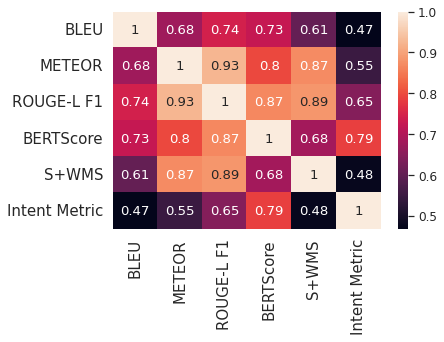} \label{corr_2}}\\
	
	\caption{Correlation Matrix for Intent Metric and other baseline metrics}
	\label{fig:corr}
\end{figure}

Also, from the correlation matrices among all the evaluation metrics corresponding to both datasets in Fig.~\ref{fig:corr}, we find that Intent Metric has the highest correlation of with S+WMS in case of `Indian Data', and with BERTScore in case of `Australian-Data'. Hence, our metric shows high correlation with metrics that quantify semantic similarity, rather than lexical similarity.

We perform our experiments on server with a RAM of $12.69$ GB and a NVIDIA Tesla K80 GPU with a $12$ GB memory.

\begin{table*}[t]
\renewcommand\thetable{5}
\centering
\resizebox{\textwidth}{!}{
\begin{tabular}{|c|cc|cc|cc|cc|cc|cc|}
\hline
\textbf{\begin{tabular}[c|]{@{}c@{}}Model Name\end{tabular}} &
  \multicolumn{2}{c|}{\textbf{BLEU}} &
  \multicolumn{2}{c|}{\textbf{METEOR}} &
  \multicolumn{2}{c|}{\textbf{\begin{tabular}[c|]{@{}c@{}}ROUGE-L F1\end{tabular}}} &
  \multicolumn{2}{c|}{\textbf{\begin{tabular}[c|]{@{}c@{}}BERT Score\end{tabular}}} &
  \multicolumn{2}{c|}{\textbf{S+WMS}} &
  \multicolumn{2}{c|}{\textbf{\begin{tabular}[c|]{@{}c@{}}Intent Metric\end{tabular}}} \\
  \hline
                                  & ID    & AD             & ID                            & AD    & ID   & AD    & ID    & AD    & ID   & AD    & ID            & AD    \\ \hline
Relevance & -0.09 & \textbf{-0.03} & -0.14 & -0.09 & 0.06 & -0.32 & 0.03  & -0.18 & 0.25 & -0.59 & \textbf{0.42} & -0.05 \\
\begin{tabular}[c|]{@{}c@{}}Human Score\end{tabular}                       & -0.02 & \textbf{0.09}  & -0.03                         & \textbf{0.09}  & 0.18 & -0.21 & -0.04 & 0.04  & 0.19 & -0.57 & \textbf{0.34} & -0.04\\ \hline
\end{tabular}
}
\caption{Spearman Rank Correlation of automated metrics with human evaluation metrics on both ID (‘Indian-Data’) and AD (‘Australian-Data’). Highest correlation corresponding to each dataset and human evaluation metric is in \textbf{bold}.}
\label{tab:metrics_corr}
\end{table*}

\subsection{Human Evaluation}
\label{humaneval}

To validate an automated evaluation metric, human evaluation of the generated summaries is necessary. We use Appen (\url{https://client.appen.com/}) 
platform to carry out the survey (\url{https://bit.ly/3n7xbCb}). 
As discussed in \cite{human_eval_1}, \cite{human_eval_2}, measuring Relevance (if the summary contains salient information from original text) and Readability (coherence and fluency of the summary) of the summaries are essential for evaluating the quality of the summary. We report Relevance and `Human Score', which is the average of Relevance and Readability.

For a survey on each dataset, $40$ documents in case of `Indian Dataset', and $20$ documents in case of `Australian Dataset' are sampled (each document has less than $20,000$ characters to reduce annotation load). These documents are randomly split into $4$ equal-sized sets, and for each set, a different summarization method is used. For each (original text, summary) pair, 3 questions are asked - (1) category of the legal case (2) Relevance (3) Readability. For Relevance and Readability, the annotator has to pick from a $1-5$ Likert Scale ('Very Poor' - 1, 'Poor' - 2, 'Fair' - 3, 'Good' - 4, 'Excellent' - 5). One document is annotated by two annotators. Average inter-annotator agreement Cohen $\kappa$ is 0.74.

\label{human_eval_section}

From Table \ref{tab:metrics_corr}, in case of `Indian-Data', Intent Metric beats other metrics in both 'Relevance' as well as 'Human Score'. In case of the `Australian-Data', the correlation of Intent Metric `Relevance' and 'Human Score' is second and third best from the highest one in both fields. However, the average correlation across the two datasets is the highest among all metrics w.r.t both Relevance ($\mathbf{0.185}$) and Human Score ($\mathbf{0.15}$). 
Hence, we can conclude that Intent Metric is an important metric in terms of overall human satisfaction.

\subsection{Working Demonstration}

This section elaborates the process by which users can implement our methods.

\begin{figure}[H]
  \includegraphics[width=\linewidth]{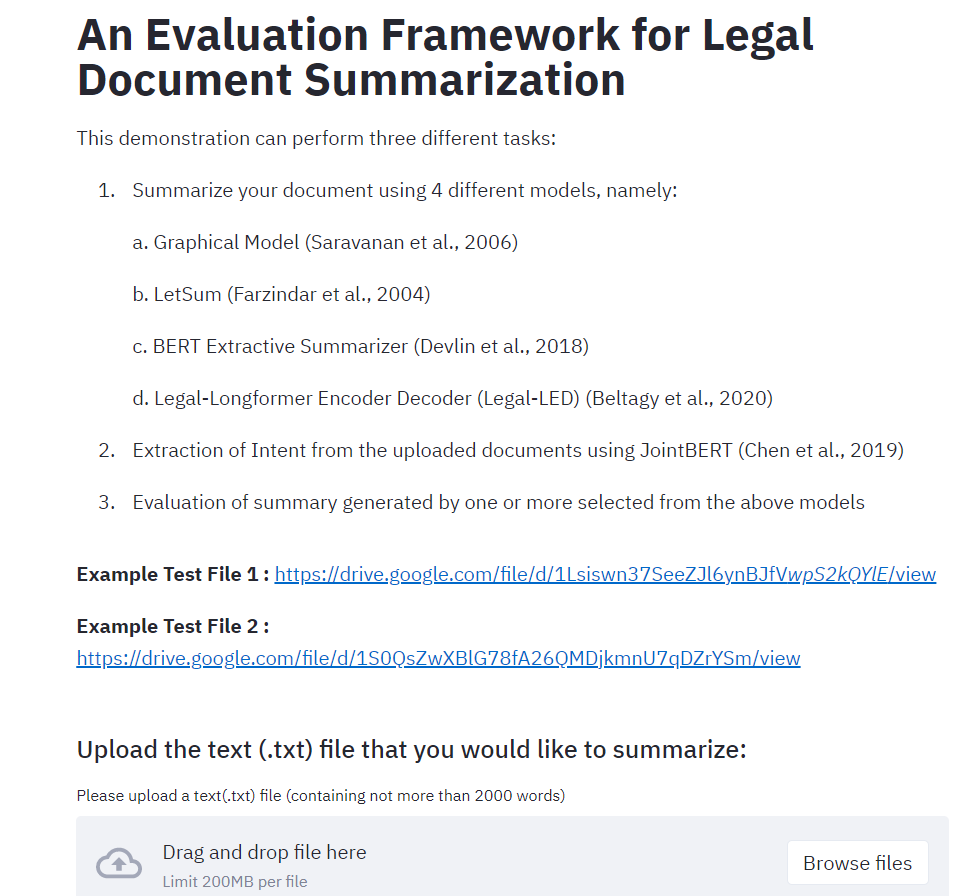}
  \caption{Landing Page of Demonstration}
  \label{fig:img1}
\end{figure}

Figure \ref{fig:img1} shows the landing page of the demonstration website. 

The demonstration can perform three different tasks:

\begin{itemize}
  \item Summarize the document you upload with either one of the four available options:
  \begin{enumerate}
  \item \textbf{Graphical Model}~\cite{graphical_model}
  \item \textbf{LetSum}~\cite{letsum}
  \item \textbf{Legal-Longformer Encoder Decoder (Legal-LED)}~\cite{legal-led}
  \item \textbf{BERT Extractive Summarizer}~\cite{bert_extractive}
  \end{enumerate}
  \item Extraction of Intent from the uploaded document data using Joint-BERT~\cite{jointbert}
  \item Evaluation of summary generated by the chosen model.
\end{itemize}

\begin{figure}[h]
  \includegraphics[width=\linewidth]{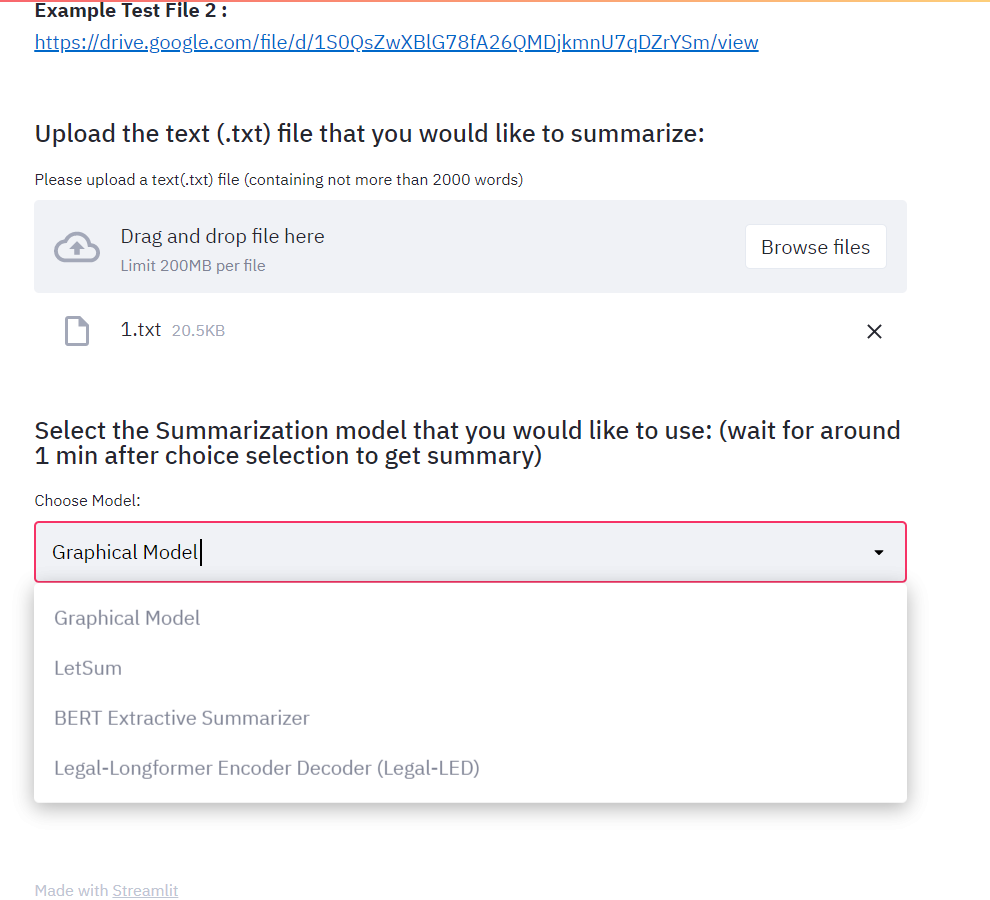}
  \caption{File upload and Model Selection Options}
  \label{fig:img2}
\end{figure}

The user can upload their own text file, or can use any one of the two example files whose link are present on the demonstration page. Furthermore, after uploading the text file, the user has to select any one of the four options available in the dropdown list of Figure \ref{fig:img2} and then select the "Click to start Summarization" option in order to run the model to start summarization of the uploaded text.

\begin{figure}[H]
  \includegraphics[width=\linewidth]{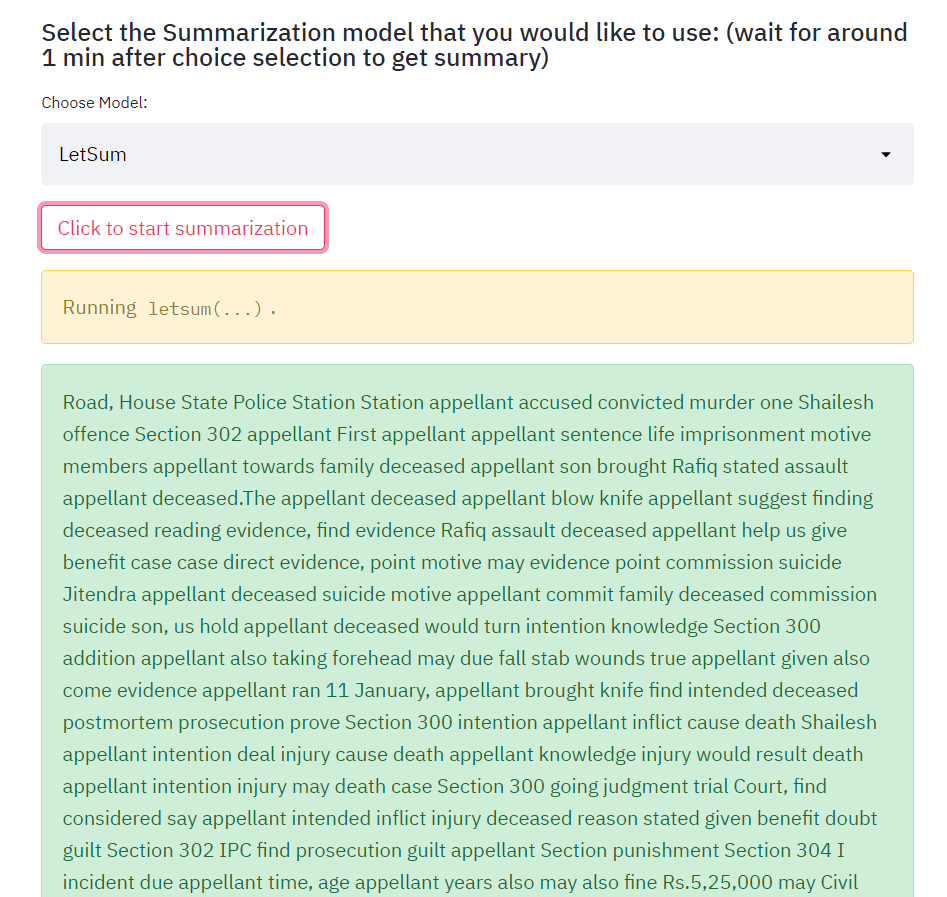}
  \caption{Summarization Output}
  \label{fig:img3}
\end{figure}

After the model is selected in Figure \ref{fig:img2}, the model is instantiated, and after sometime output summary is shown as output in the green box as seen in Figure \ref{fig:img3}.

\begin{figure}[H]
  \includegraphics[width=\linewidth]{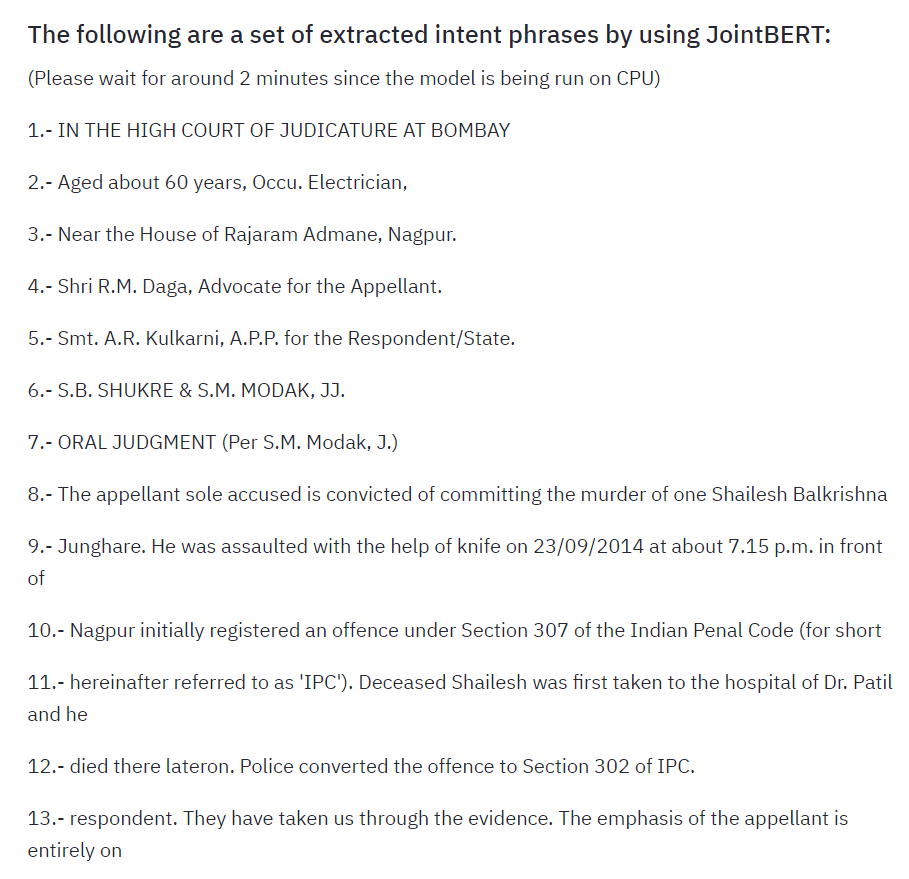}
  \caption{Extracted Intent Phrases from JointBERT}
  \label{fig:img5}
\end{figure}

\begin{figure}[H]
  \includegraphics[width=\linewidth]{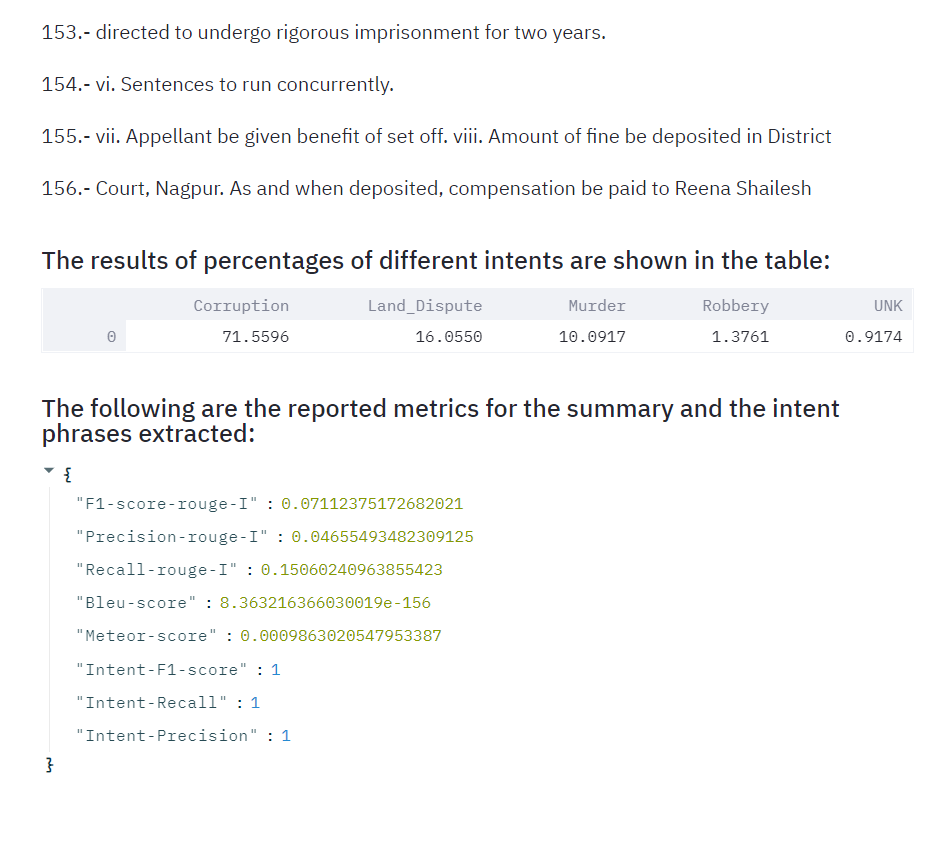}
  \caption{Intent Classification and Evaluation Results}
  \label{fig:img6}
\end{figure}

Once, the data is summarized using the model of user's choice. The demonstration automatically instantiates JointBERT \cite{jointbert} for automated intent phrase extraction from the original data as seen in Figure \ref{fig:img5} for further evaluation using our proposed Intent Metric. Furthermore, JointBERT also performs intent classification which gives the percentage of each intent present in the uploaded document whose output could be seen in Figure \ref{fig:img6}.

\section{Conclusion}

In this paper, we explore a far less studied problem of devising a suitable evaluation metric for legal document summarization. To tackle the problem, a pre-condition is to curate a dataset that contains intent phrases extracted from legal documents belonging to categories like Robbery, Land Dispute, etc. This helps to develop a metric that correlates with human readability and relevance comparatively more than other metrics. We show a proof of concept that such intent phrase annotations required for the calculation of Intent Metric can be automated (Australian-Data). We believe that, such a metric would serve a better purpose in evaluating summarization of legal documents. We plan to extend the work on different categories of legal documents for various countries. We shall make the code, data available after acceptance.
\section*{References}\label{reference}
\bibliographystyle{lrec2022-bib}
\bibliography{custom}

\end{document}